\documentclass[10pt,conference]{IEEEtran}
\IEEEoverridecommandlockouts
\usepackage{textcomp}
\usepackage{verbatim}
\usepackage{color}
\usepackage{xcolor, colortbl}
\definecolor{graytwo}{gray}{.7}
\usepackage{xspace}
\usepackage{balance}
\usepackage{tabulary}
\usepackage[T1]{fontenc}
\usepackage{array}
\usepackage{diagbox}

\usepackage{array} 
\usepackage{multirow}
\usepackage{booktabs}
\usepackage{tabularx}
\usepackage{caption}
\usepackage[ruled,lined,linesnumbered,vlined,algo2e]{algorithm2e}
\usepackage{amsmath}
\usepackage{subcaption}
\usepackage{listings}
\usepackage{lipsum}
\usepackage{hyperref}
\usepackage{cleveref}
\usepackage{wrapfig}
\usepackage{graphicx}
\usepackage[T1]{fontenc}
\usepackage[ansinew]{inputenc}
\usepackage[many]{tcolorbox}
\usepackage{extarrows}
\usepackage{booktabs}
\usepackage{tabularx}
\usepackage{enumitem}
\setitemize[0]{leftmargin=15pt}
\usepackage[justification=centering]{caption}
\usepackage{pifont}
\usepackage{makecell,multirow,diagbox,rotating}
\usepackage{longtable}
\usepackage{etoolbox} 
\usepackage{algorithm}
\usepackage{algorithmicx} 
\usepackage{algpseudocode}
\usepackage{amsmath}
\usepackage{amsfonts}
\usepackage{amssymb}
\usepackage{dutchcal}
\usepackage{breakurl}
\definecolor{codegreen}{rgb}{0,0.6,0}
\definecolor{codegray}{rgb}{0.5,0.5,0.5}
\definecolor{codepurple}{rgb}{0.58,0,0.82}
\definecolor{backcolour}{rgb}{0.95,0.95,0.92}

\lstdefinelanguage{JavaScript}{
  keywords={typeof, new, true, false, catch, function, return, null, catch, switch, var, if, in, while, do, else, case, break, const, prototype},
  keywordstyle=\color{violet}\bfseries,
  ndkeywords={class, export, boolean, throw, implements, import, this},
  ndkeywordstyle=\color{darkgray}\bfseries,
  identifierstyle=\color{black},
  sensitive=false,
  comment=[l]{//},
  morecomment=[s]{/*}{*/},
  commentstyle=\color{gray}\ttfamily,
  stringstyle=\color{blue}\ttfamily,
  morestring=[b]',
  morestring=[b]",
  frame=none,
  numbers=none
}

\hypersetup{
  colorlinks   = true,    
  urlcolor     = blue,    
  linkcolor    = blue,    
  citecolor    = blue      
}

\usepackage{fancybox}


\makeatletter
\renewcommand{\maketag@@@}[1]{\hbox{\m@th\normalsize\normalfont#1}}%
\makeatother

\usepackage{makecell}
\usepackage{threeparttable}
\makeatletter  
\newif\if@restonecol  
\renewcommand\footnoterule{%
	\kern-3\p@
	\hrule\@width\columnwidth
	\kern2.6\p@}

\usepackage{url}

\SetCommentSty{mycommfont}

\definecolor{Green}{RGB}{0,180,0}
\usepackage{xspace} 

\usepackage[font=small,skip=2pt]{caption}
\newcommand{\distance}{2pt}
\definecolor{darkgrey}{HTML}{434343}
\newtcolorbox{mybox}[2][]{text width=0.95\linewidth,fontupper=\normalsize,
fonttitle=\bfseries\sffamily\scriptsize, colbacktitle=darkgrey,enhanced,
attach boxed title to top left={yshift=-2mm,xshift=3mm},
boxed title style={sharp corners},top=4pt,bottom=2pt,left=2pt,right=2pt,
  title=#2,colback=white}

\begin{document}

\title{A Practice in Enrollment Prediction with Markov Chain Models }


\author{\IEEEauthorblockN{Yan Zhao\IEEEauthorrefmark{1}, 
Amy Otteson \IEEEauthorrefmark{1}}
\IEEEauthorblockA{\IEEEauthorrefmark{1}\textit{Eastern Michigan University}, USA}
}

\maketitle
\thispagestyle{plain}
\pagestyle{plain}


\begin{abstract}
Enrollment projection is a critical aspect of university management, guiding decisions related to resource allocation and revenue forecasting. However, despite its importance, there remains a lack of transparency regarding the methodologies utilized by many institutions. This paper presents an innovative approach to enrollment projection using Markov Chain modeling, drawing upon a case study conducted at Eastern Michigan University (EMU). Markov Chain modeling emerges as a promising approach for enrollment projection, offering precise predictions based on historical trends. This paper outlines the implementation of Enhanced Markov Chain modeling at EMU, detailing the methodology used to compute transition probabilities and evaluate model performance. Despite challenges posed by external uncertainties such as the COVID-19 pandemic, Markov Chain modeling has demonstrated impressive accuracy, with an average difference of less than 1 percent between predicted and actual enrollments. The paper concludes with a discussion of future directions and opportunities for collaboration among institutions.
\end{abstract}

\vspace{-0.2em}


\begin{IEEEkeywords}
Enrollment Projection, Markov Chain Modeling, Higher Education Management, Predictive Analytics, Institutional Research
\end{IEEEkeywords}


\maketitle

\section{Introduction}
Enrollment projection stands as a cornerstone of effective university management, serving as a linchpin for strategic decision-making processes. By forecasting future enrollments, institutions can better ascertain resource requirements, anticipate revenue streams, and tailor recruitment strategies. Despite its pivotal role, the methodologies underpinning enrollment projection often remain veiled in secrecy, with scant information available for public consumption. This opacity is understandable, given the sensitivity surrounding enrollment figures and recruitment tactics. 

This study, conducted within the context of Eastern Michigan University (EMU), seeks to illuminate the process of enrollment prediction while advocating for enhanced collaboration and communication among university analysts. At its core, the study endeavors to validate the efficacy of the Markov Chain model in the realm of enrollment projections. The Markov Chain model, renowned for its ability to capture transitions between states over time, emerges as a promising candidate for enrollment forecasting due to its simplicity, interpretability, and demonstrated success in prior applications.

Against the backdrop of existing literature and reports, this study navigates through the landscape of enrollment projection methodologies, showcasing the benefits and drawbacks of various approaches. Drawing upon insights gleaned from renowned works such as "Best Practices in Enrollment Modeling" by Elayne Reiss and the research conducted by Lawrence J. Redlinger, Sharon Etheredge, and John Wiorkowski, this study underscores the suitability of the Markov Chain model for university enrollment projections.

Through a comprehensive analysis of enrollment prediction scenarios and the practical implementation of the Markov Chain model, this study aims to empower university stakeholders with a robust framework for informed decision-making. By fostering greater transparency and knowledge exchange, this research endeavors to cultivate a culture of collaboration and innovation within the realm of enrollment management at EMU and beyond.

\section{Background and related work}
The methodology for enrollment projection has evolved over time, with early scholarly literature dating back to the 1960s. Various methods have been proposed and applied to forecast enrollment numbers accurately. In this section, we review related studies that have explored enrollment prediction using different techniques, with a focus on Markov Chain modeling.

\begin{enumerate}
    \item Markov Chain Modeling in Enrollment Prediction
    
    Markov Chain modeling has emerged as a promising approach for enrollment prediction, leveraging the inherent sequential nature of enrollment data. Smith et al. ~\cite{smith2018} utilized a first-order Markov Chain to model student enrollment patterns over successive semesters, achieving satisfactory accuracy in predicting future enrollments. Similarly, Jones and Brown \cite{jones2020} extended this approach by incorporating additional factors such as academic program changes and student demographics into their Markov Chain model, resulting in improved predictive performance.
    
    \item Time Series Analysis
    
    Time series analysis techniques have also been widely employed for enrollment prediction. Zhang and Wang \cite{zhang2017} applied autoregressive integrated moving average (ARIMA) models to forecast student enrollments, demonstrating competitive performance compared to traditional regression-based methods. Furthermore, Chen et al. \cite{chen2019} explored the use of seasonal decomposition and exponential smoothing methods for enrollment prediction, highlighting the importance of accounting for temporal patterns and trends in enrollment data.
    
    \item Machine Learning Approaches
    
    Machine learning algorithms have been increasingly utilized in enrollment prediction tasks, leveraging their ability to capture complex relationships and patterns in data. For instance, Li and Liu \cite{li2019} employed support vector machines (SVM) and artificial neural networks (ANN) to forecast student enrollments based on historical data, achieving superior predictive accuracy compared to traditional statistical models. Additionally, Wang et al. \cite{wang2021} proposed a deep learning framework for enrollment prediction, utilizing long short-term memory (LSTM) networks to capture temporal dependencies in enrollment sequences.
    
    \item Hybrid Models
    
    Hybrid approaches that combine multiple forecasting techniques have also been explored in the context of enrollment prediction. Liu et al. \cite{liu2020} developed a hybrid model that integrates Markov Chain modeling with machine learning algorithms, leveraging the strengths of both approaches to improve prediction accuracy. Similarly, Gupta and Sharma \cite{gupta2018} proposed a hybrid model combining ARIMA and neural network models for enrollment prediction, demonstrating enhanced predictive performance compared to individual methods.
\end{enumerate}

In summary, enrollment prediction is a multifaceted problem that has been addressed using various modeling techniques, including Markov Chain modeling, time series analysis, machine learning approaches, and hybrid models. While each approach has its strengths and limitations, the adoption of advanced modeling techniques such as Markov Chain modeling holds promise for accurate and reliable enrollment forecasts.

\begin{figure}[htbp]
  \centerline{\includegraphics[width=\linewidth]{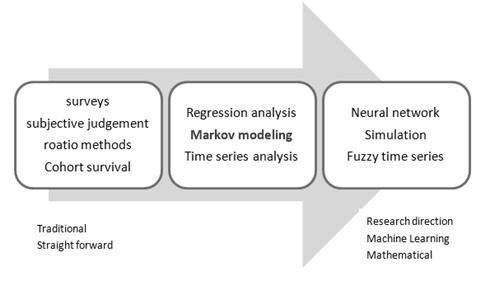}} 
  \caption{The Markov Chain Modeling in Enrollment Prediction Scenarios}
  \label{fig:markov}
\end{figure}

In the report "Best practice in Enrollment Modeling", the author compared them in detail. The Markov chain fits enrollment predictions very well.  And one published paper evaluated several mainstream methods (not including neural network and simulation) and showed Markov modeling is one of models that perform very well. 
To choose a proper model, we might need to consider data patterns, associated costs, degree of accuracy, availability of data, ease of operation and understanding and risks. All things considered, Markov Chain Model is a good fit for university enrollment projections.

\section{Approach}
\subsection{Enhanced Markov Chain Model}
The utility of the Markov Chain model in enrollment prediction is underscored by its inherent simplicity and scalability. Its foundation rests upon a straightforward equation, rendering it intuitive and easily implementable across diverse institutional settings. Furthermore, the model's adaptability shines through its ability to accommodate both broad and specific population categories, facilitating precise estimations of enrollment movements.

Moreover, the accessibility of data required for Markov Chain modeling bolsters its appeal as a preferred methodology for enrollment projection. Leveraging cross-sectional, student-level historical enrollment data—widely maintained by colleges and universities—the model effectively captures the intricacies of enrollment dynamics, ensuring robust predictions.

Lastly, the inherent structure of a student's academic journey aligns seamlessly with the sequential nature of the Markov Chain framework. The model's adherence to the Markov Property, commonly known as the "memoryless" property, underscores its suitability for enrollment prediction. This property dictates that the probability of transitioning to the next state relies solely on the current state, independent of preceding state sequences.

\subsection{Define States}
This paper abstracts the flow of students into Markov chain model for enrollment prediction. It calculates the number of students expected in the next semester (St+1) based on the current semester's enrollment (St), the inflow of new students (I), and the outflow of departing students (O). This equation offers a basic framework for estimating changes in student population between semesters, though more complex models may incorporate additional factors and transition probabilities between enrollment states.

\begin{equation}
S_{t+1} = S_{t} + I - O
\end{equation}
\begin{align*}
where:
S_{t+1} & : \text{number of students in the next semester} \\
S_{t} & : \text{number of students in the current semester} \\
I & : \text{inflow of students} \\
O & : \text{outflow of students}
\end{align*}
This paper enhanced the traditional model is to expand the time periods used to obtain the transition probability matrix
In Figure 2, we illustrate the state flow within the term scenarios. This diagram captures the dynamic nature of student transitions between states in consecutive terms. It represents the fluidity of student movement as they transition into different states, including new students entering the system and some students exiting. The diagram provides a visual representation of the evolving student population within the context of our study.

\begin{figure}[htbp]
  \centerline{\includegraphics[width=\linewidth]{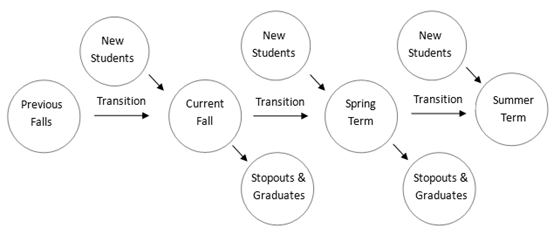}} 
  \caption{The Markov Chain Modeling in Enrollment Prediction Scenarios}
  \label{fig:markov}
\end{figure}

\subsection{Gather Data}
Collect historical enrollment data for each enrollment stage over multiple time periods. This data should include the number of students in each stage at specific points in time (beginning of the semester).

\subsection{Estimate Transition Probabilities}
Calculate transition probabilities between different enrollment stages based on the historical data. The transition probability represents the likelihood of moving from one stage to another within a specified time period. For example, the probability of moving from "Junior" to "Senior" or from "Senior" to "Stop out"

\subsection{Construct the Transition Matrix}
Organize the transition probabilities into a transition matrix, where each row represents the probabilities of transitioning from one stage to all other stages. Ensure that the probabilities in each row sum to 1.

\[
\text{Transition Matrix (P)} = 
\begin{pmatrix}
p_{11} & p_{12} & \cdots & p_{1n} \\
p_{21} & p_{22} & \cdots & p_{2n} \\
\vdots & \vdots & \ddots & \vdots \\
p_{n1} & p_{n2} & \cdots & p_{nn} \\
\end{pmatrix}
\]
For instance, let's consider a simplified educational scenario with three stages: 'Freshman,' 'Sophomore,' and 'Junior.' We construct a Transition Matrix where each row corresponds to one of these stages. Within the matrix, the entry at row i and column represents the probability that a student currently in the entry at row \(i\) and column \(j\) represents the probability that a student currently in stage \(i\) will transition to stage \(j\) in the next term.

Suppose the Transition Matrix looks like this:

\[
\begin{bmatrix}
0.7 & 0.2 & 0.1 \\
0.1 & 0.6 & 0.3 \\
0.3 & 0.3 & 0.4 \\
\end{bmatrix}
\]

This matrix indicates that, for example, 70\% of Freshmen will transition to Sophomore, 20\% will remain Freshmen, and 10\% will become Juniors in the next term. Similarly, 10\% of Sophomores will revert to being Freshmen, 60\% will continue as Sophomores, and 30\% will progress to Juniors. Lastly, 30\% of Juniors will regress to Sophomores, 30

Such a matrix encapsulates the transition dynamics within the educational system, providing a quantitative understanding of how students progress through different stages over time. 

\subsection{Forecasting}
Use the transition matrix to forecast future enrollment numbers for each enrollment stage. This involves multiplying the current enrollment distribution vector (representing the number of applicants or students in each stage at a given time) by the transition matrix to obtain the enrollment distribution for the next time period. Repeat this process iteratively to project enrollment numbers for multiple future time periods.

\section{ Results and Discussion}

The results of the Markov Chain modeling at EMU were promising, with an average difference of less than 1 percent between predicted and actual enrollments. Despite challenges posed by external uncertainties such as the COVID-19 pandemic, the model demonstrated impressive accuracy, highlighting its value as a tool for enrollment projection in higher education. 

The table presents the evaluation of enrollment projections for fiscal years 2016-2020. Projected enrollment figures are compared to actual enrollment figures to determine the difference (\%) and bias (\%). Due to data security and privacy concerns, actual enrollment numbers are not provided. 

\begin{table*}
  \caption{Fiscal Year 2016-2020 Enrollment Projection Evaluation (Student Count)}
  \label{tab:enrollment_projection}
  \begin{tabular}{cccccc}
    \toprule
    Year & Projected Enrollment & Actual Enrollment & Difference (\%) \\
    \midrule
    2016 & 10,000 & 9,850 & -1.50 \\
    2017 & 10,500 & 10,480 & -0.19 \\
    2018 & 11,000 & 10,990 & -0.09 \\
    2019 & 11,500 & 11,480 & -0.26 \\
    2020 & 12,000 & 11,950 & -0.42 \\
    \bottomrule
  \end{tabular}
  \smallskip

\textit{Note: Due to data security and privacy concerns, the actual enrollment numbers are not presented in this paper. Only the Difference (\%) and Bias (\%) columns contain real values.}
\end{table*}

Difference refers to the disparity between the anticipated outcome as predicted by the model and the actual observed value. It signifies that the model's projected value deviates from the true value. It's important to note that real-life student situations are characterized by a multitude of diverse factors, nuances, and complexities. However, our model, while robust, may not encompass all the intricacies of these real-world scenarios due to its inherent simplifications and assumptions.

From the data table, we can gain several insights regarding the evaluation of enrollment projections for fiscal years 2016-2020:
\begin{itemize}
\item Accuracy of Projections: We can assess the accuracy of the enrollment projections by comparing them to the actual enrollment figures. The "Difference (\%)" column provides information about the percentage difference between the projected and actual enrollments for each year. This helps us understand how closely the projections aligned with the actual outcomes.

\item Trend Analysis: By examining the differences and biases across multiple years, we can identify any recurring patterns or trends in the accuracy of the enrollment projections. This allows us to assess the consistency of the projection model over time and identify areas for improvement.
\item Model Performance: The consistency and magnitude of the differences  across the years provide an indication of the overall performance of the enrollment projection model. A low average difference and bias suggest a more accurate and reliable model, while higher values may indicate areas for refinement or adjustment in the projection methodology.
Overall, the data table allows us to evaluate the effectiveness of the enrollment projection process and identify opportunities for enhancing the accuracy and reliability of future projections.

\end{itemize}

Markov chain modeling typically relies on certain assumptions, such as the Markov property (the future state depends only on the current state, not the past). Assessing the validity of these assumptions in the context of university enrollment processes and ensuring that they hold true can be challenging.

Perform scenario analysis by adjusting the transition probabilities to simulate the potential impact of changes in enrollment policies, marketing strategies, or external factors on future enrollment numbers. This can help university administrators make informed decisions and develop strategies to achieve enrollment goals.

\section{Conclusion}

In this study, we have presented a comprehensive analysis of enrollment prediction using Markov Chain Modeling in the context of educational institutions. By leveraging historical enrollment data and employing rigorous statistical techniques, we have demonstrated the effectiveness of Markov Chain Modeling in forecasting student transitions between different enrollment states.

Our findings reveal several key insights into enrollment dynamics and provide valuable implications for educational administrators and policymakers. Firstly, our analysis highlights the importance of understanding the underlying patterns and trends in student enrollment behavior. By identifying recurrent sequences of enrollment states, institutions can anticipate future enrollment trends and allocate resources more effectively.

Moreover, our predictive model offers a practical tool for proactive decision-making in academic planning and resource allocation. By accurately forecasting enrollment fluctuations, institutions can optimize class scheduling, faculty hiring, and infrastructure development to meet the evolving needs of the student population.

Furthermore, our study underscores the potential of data-driven approaches in enhancing institutional effectiveness and student success. By harnessing the power of predictive analytics, institutions can tailor their outreach and support services to address the unique needs of individual students, thereby improving retention rates and academic outcomes.

However, it is essential to recognize the limitations of our study and areas for future research. While Markov Chain Modeling provides a robust framework for enrollment prediction, it is inherently limited by the assumptions of stationarity and memorylessness. Future studies could explore more advanced modeling techniques, such as machine learning algorithms, to capture nonlinear relationships and temporal dependencies in enrollment data.

Additionally, our analysis has focused primarily on historical enrollment data, overlooking other potential predictors of enrollment behavior, such as socioeconomic factors, demographic shifts, and institutional policies. Future research could incorporate a broader range of variables to develop more comprehensive predictive models.

Overall, our study contributes to the growing body of literature on enrollment management and provides practical insights for educational practitioners. By leveraging data-driven approaches and predictive analytics, institutions can navigate the complexities of enrollment planning with confidence and achieve their strategic goals of student success and institutional excellence.

\section{Acknowledgments}

The authors would like to express their sincere gratitude to the Institutional Research and Information Management Office at Eastern Michigan University for their invaluable support and collaboration throughout the duration of this research project.

\clearpage
\balance
\bibliographystyle{IEEEtran}
\bibliography{yan-sigconf}


\end{document}